\documentclass[11pt]{article}
\usepackage[final]{acl}
\usepackage{times}
\usepackage{latexsym}
\usepackage[T1]{fontenc}
\usepackage[utf8]{inputenc}
\usepackage{microtype}
\usepackage{inconsolata}
\usepackage{array}
\usepackage{booktabs}
\usepackage{multirow}
\usepackage{amsmath}
\usepackage{amssymb}
\usepackage{graphicx}

\title{Acquire, Repair, Preserve: A Diagnosis-Guided Post-Training Recipe for Small-Model Dialogue Game Agents}

\author{Nan Li \\
        Department of Information and Computing Sciences,\\Utrecht University, Utrecht, The Netherlands\\
        \texttt{n.li@uu.nl}}

\begin{document}
\maketitle

\begin{abstract}
Interactive dialogue games test a capability that static benchmarks largely leave implicit: a model must carry state across turns, interpret feedback, and choose valid actions under changing constraints.
We study this setting in the LM Playschool Challenge with a 2B open-weight model, and find that many failures are not only broad knowledge failures but also local decision failures: repeated guesses, malformed actions, and violations of feedback that the model has just seen.
These diagnostics motivate a training recipe organized around three steps: acquire broad game participation through supervised fine-tuning, repair mechanically verifiable failures within one targeted dialogue-game family using turn-local preference pairs, and preserve general capabilities beyond these dialogue games.
In the official final evaluation, our submission improves public clemscore from 10.67 to 38.92 and closed in-domain score from 13.41 to 41.17, while approximately preserving aggregate static performance (44.14 vs.\ 44.24 for the baseline).
Out-of-domain clemscore remains low at 7.88, with the largest gains concentrated in unseen variants of the targeted family.
Our results suggest that broad SFT brings most of the model's capability improvement; turn-local supervision can be effective when failure detection is precise, with observed transfer concentrated primarily within-family.
\end{abstract}

\section{Introduction}

Dialogue games evaluate language models as agents whose next utterance changes the state they must later reason over.
A word guess rules out letters; a clue constrains a partner's action; a navigation command changes the room from which the next command is interpreted.
The clembench benchmark and the Playpen learning environment \citep{chalamalasetti2023clembench,playpen2025} make these dependencies explicit: success requires not only producing plausible language, but also maintaining state, interpreting feedback, and choosing a valid action under accumulated constraints.

Playpen frames this setting as learning from Dialogue Game Feedback~\citep{playpen2025}: programmatic, goal-directed feedback over trajectories, without relying on a learned judge over isolated responses.
The feedback does not itself locate where an interaction stops making progress; interactive failure diagnostics are needed to construct turn-local preference pairs for reinforcement learning. 
The question for a dialogue game agent is therefore not only whether dialogue game data helps, but which part of the trajectory should receive supervision.

Small models are a natural target for this setting. They are cheaper to serve, better suited to low-latency interaction, and more plausible for local assistants or assistants with limited resources, but their interactive behavior is brittle. In the LM Playschool Challenge, the official \mbox{Qwen3.5-2B} baseline~\citep{qwen2026qwen35} scores 10.67 clemscore in the final evaluation, far below the 60.30 score of its 27B sibling.\footnote{Official final evaluation results: \url{https://github.com/lm-playpen/lm-playschool-2026-final-results}.} The challenge also reports a static benchmark score, so a useful small model recipe must improve interaction without sacrificing general capability.

\begin{table}[t]
\centering\small
\setlength{\tabcolsep}{3pt}
\begin{tabular}{@{}p{.42\linewidth}p{.27\linewidth}p{.23\linewidth}@{}}
\toprule
Dialogue state & Model action & Local failure \\
\midrule
After \texttt{crane}: red \texttt{c/r/n}; yellow \texttt{a}; green \texttt{e}
& \texttt{guess: crane}
& repeats guess \\
\quad\textit{(same)}
& \texttt{guess: abid}
& wrong length \\
\quad\textit{(same)}
& \texttt{guess: cable}
& uses red \texttt{c} \\
\bottomrule
\end{tabular}
\caption{Simplified Wordle failure states. The model output can be syntactically close to valid while violating a constraint entailed by the immediately preceding feedback. A real transcript exhibiting this pattern is reproduced in Table~\ref{tab:wordle-real}.}
\label{tab:wordle-example}
\end{table}

Inspecting the 2B model's transcripts suggested that the gap was not mainly a lack of game knowledge. In Wordle, the model often preserves the expected answer format while failing at the next local decision (Table~\ref{tab:wordle-example}). These errors are frequent, mechanically checkable, and localized to a specific decision made after receiving feedback. They point to a different training question: not how to teach the model the whole dialogue again, but where to place supervision so that it reaches the particular decision that fails.

This paper describes the \texttt{playornotplay} team submission built around that question. We first use broad imitation to make the model participate in games, then diagnose the remaining failures, repair those of one weak and mechanically diagnosable game family (Wordle) with turn-local preference pairs, and finally scale the learned update to preserve general capabilities, as measured by aggregate static performance.
In the official final evaluation, the submitted model reaches 38.92 public clemscore and 41.17 closed in-domain score, above the official 4B and 9B baselines, while approximately preserving aggregate static performance over the 2B baseline (44.14 vs.\ 44.24).

The main contribution is an empirical account of supervision placement for small interactive agents.
A small number of preference pairs anchored to real turns after feedback produces the clearest positive preference-tuning signal in our pipeline, whereas whole-dialogue preference tuning destroyed protocol compliance in our experiments.
Follow-up experiments further show that the usefulness of an update depends on the stage and parent to which it is applied. Together, these results locate turn-local preference learning within a broader progression from acquisition to repair and model selection.

\section{Task and Evaluation Setting}
\label{sec:eval}

The LM Playschool Challenge evaluates whether learning in conversational interaction improves agentic capability. Playpen is the learning environment for Dialogue Game Feedback \citep{playpen2025}: it organizes rollouts, training data, and online sampling over dialogue game trajectories. The dialogue games are implemented in clembench \citep{chalamalasetti2023clembench}, where programmatic game masters mediate turns, track state, and enforce output protocols, while game-specific scorers compute Quality.

In the public validation setting used here (from the \texttt{playpen-data} dataset~\footnote{\url{https://huggingface.co/datasets/colab-potsdam/playpen-data}.}), the interactive suite contains 14 clembench games and 67 episodes in our pinned development run.
We pin the public-development data to revision \texttt{557d8caf} and the clembench game tree to commit \texttt{ed39486}.
The official final evaluation comprises an expanded public suite (502 episodes), a closed in-domain suite (1{,}272 episodes), and a closed out-of-domain suite covering unseen game families (360 episodes)~\footnote{\url{https://github.com/lm-playpen/lm-playschool-2026-closed-set}}.
We use the public-development set for iteration and reserve the closed suites for final evaluation.

The LM Playschool Challenge uses Clemscore and Statscore as metrics. \textbf{Clemscore} combines two quantities for each game $g$: \textbf{Played}, the percentage of episodes not marked aborted by the evaluator, and \textbf{Quality}, the task score on played episodes. With $G_Q=\{g\in G:\mathrm{Played}_g>0\}$, we report it as
\begin{equation}
\begin{split}
C={}&\left(\frac{1}{|G|}\sum_{g\in G}\frac{\mathrm{Played}_g}{100}\right)\\
&\times\left(\frac{1}{|G_Q|}\sum_{g\in G_Q}\mathrm{Quality}_g\right).
\end{split}
\label{eq:clemscore}
\end{equation}
Quality is undefined for a game with no played episode, so the Quality macro-average runs over $G_Q$ only. Thus a model can lose score either by failing to play or by playing poorly after it stays in the game. The \texttt{playpen-data} validation split contains 430 static-benchmark instances drawn from BBH \citep{suzgun2023bbh}, CLadder \citep{jin2023cladder}, EQ-Bench \citep{paech2023eqbench}, IFEval \citep{zhou2023ifeval}, and MMLU-Pro \citep{wang2024mmlupro}; applying the same aggregation yields \textbf{Statscore}.

The pinned public split contributes one or two episodes for each represented game--experiment combination, and this small sample is the main source of run-to-run variance: differences of a few points should be treated cautiously. We therefore use paired same-host comparisons for development decisions and report the official final evaluation as the primary result.

\section{Our Method}
\label{sec:method}

Our recipe is a narrow instantiation of the training regimes enabled by Playpen: successful transcripts support SFT, paired alternatives support DPO, and fresh rollouts support online reward learning. Playpen itself includes turn-level DPO~\citep{playpen2025}; our specific choice is to use game-specific checks to construct preference pairs at mechanically verifiable failures after feedback. We use broad SFT to make the model participate, then target those diagnosed decisions.

All training uses LoRA~\citep{hu2022lora} ($r{=}16$, $\alpha{=}32$, all-linear, adapter dropout 0.05) on \mbox{Qwen3.5-2B}.\footnote{Qwen3.5-2B~\citep{qwen2026qwen35} is a vision-language model; we fine-tune and serve it in text-only mode.}
Source trajectories and instances come from the \texttt{train} split of \texttt{playpen-data}; Phase~B negatives are synthetic corruptions using the clembench word list; Phase~C data is model-generated. Table~\ref{tab:trajectory} gives the trajectory on the pinned public-development set. 
Following the official clemscore definition in Eq.~\ref{eq:clemscore}, broad imitation primarily improves Played; our later phases target Quality on states the model already plays; final delta scaling targets statscore rather than game behavior directly.\footnote{The manuscript submitted for review described the training process for the model, which is also used in the official final evaluation, as six stages. Subsequent verification showed that two nominal stages did not update the model, so we omit them here. Appendix~\ref{app:artifact} identifies the evaluated model revision; Appendix~\ref{app:noop} explains the two removed stages.}

\begin{table}[tbp]
\centering\small
\setlength{\tabcolsep}{2pt}
\begin{tabular}{@{}lllr@{}}
\toprule
Phase & Objective & LR / batch / len & Clem \\
\midrule
--- & Qwen3.5-2B & --- & 13.05 \\
\midrule
A.\ Acquire & broad SFT & 2e\nobreakdash-4\,/\,32\,/\,2048 & 43.85 \\
\midrule
\multirow{2}{*}{B.\ Repair} & turn-DPO: rpt + len & 5e\nobreakdash-6\,/\,16\,/\,1024 & 46.83 \\
 & turn-DPO: inv + dup & 5e\nobreakdash-6\,/\,16\,/\,1024 & 45.52 \\
\midrule
C.\ Refine & branch-DPO & 5e\nobreakdash-6\,/\,16\,/\,1024 & 48.52 \\
\midrule
\multirow{2}{*}{D.\ Preserve} & delta scaling, $s{=}0.85$ & no gradient & 50.43$^\dagger$ \\
 & \emph{shipped fp32 merge} & --- & 48.70 \\
\bottomrule
\end{tabular}
\caption{Pipeline trajectory on the pinned public-development set (dataset \texttt{557d8caf}, clembench \texttt{ed39486}). Batch denotes effective batch size; all Clem values are the full 14-game suite. Phases~B and~C train on Wordle data only. Both Phase~B passes use $\beta{=}0.1$; Phase~C uses $\beta{=}0.2$. All trainable phases use LoRA on \mbox{Qwen3.5-2B}, AdamW, bf16, and one epoch; Phase~C branches the model (factor~2, rounds~1--3, $T{=}0.7$) against a frozen reference. rpt/len = repeated guesses and bad length; inv/dup = invalid words and duplicated answer fields. $^\dagger$Mean of two runs (50.82 and 50.03). Per-game and per-phase Wordle detail: Tables~\ref{tab:app-pergame} and~\ref{tab:app-wordle}.}
\label{tab:trajectory}
\end{table}

\subsection{Diagnosing Small-Model Game Failures}

Phase~A performs SFT on success-only interactions from the training split of \texttt{playpen-data}, balanced to at most 700 examples per game (9{,}522 rows total at the reconstructed training-time snapshot; per-game counts in Table~\ref{tab:sft-games}). This lifts the model from 13.05 to 43.85 on the pinned public-development set. The model has learned the surface form of game play, yet raw Wordle quality remains at zero: it still repeats rejected guesses and violates immediate feedback constraints. Wordle was not the only remaining weakness, but its failures were frequent, local to a single decision, and mechanically verifiable, so we selected it as the target of all subsequent optimization. Phases~B and~C use only Wordle data; every reported clemscore remains the full 14-game suite.

\subsection{Turn-Local Repair}

Given a diagnosed failure mode, we construct preference pairs that share the same dialogue state. For a history $h$, the rejected completion $y^-$ exhibits one mechanical error and the preferred completion $y^+$ is a valid alternative from a clean successful episode. We then apply Direct Preference Optimization (DPO;~\citealp{rafailov2023dpo}):
\begin{equation}
\mathcal{L}_{\mathrm{DPO}}=-\mathbb{E}\log\sigma\!\Big(\beta\big[r_\theta(h,y^+)-r_\theta(h,y^-)\big]\Big),
\label{eq:dpo}
\end{equation}
with policy--reference log-ratio $r_\theta(h,y)=\log\frac{\pi_\theta(y\mid h)}{\pi_{\mathrm{ref}}(y\mid h)}$. Holding $h$ fixed is the point: the contrast lands on the few decision tokens that determine whether the model repeats a guess, uses the wrong length, produces an invalid word, or duplicates the answer field. This construction does not identify a single failure; it enumerates all eligible post-feedback turns in clean successful episodes and creates a synthetic negative at each one.
Appendix~\ref{app:turn-dpo} provides the source-episode filters, turn-eligibility criteria, and the four corruption rules.

Phase~B applies this in two passes: the first addresses repeated guesses and bad length (1{,}905 pairs, $\beta{=}0.1$); the second addresses invalid words and duplicated answer fields (1{,}906 pairs, $\beta{=}0.1$). After the first pass, the old repeat loops no longer appear in the raw Wordle probe, but invalid words and duplicated answer fields emerge instead. The second pass corrects these; its aggregate suite effect is sign-inconsistent across the two evaluation rails (Table~\ref{tab:ablation}).

After addressing fixed failure classes with synthetic pairs, we turn to errors arising in the current model's own trajectories. Phase~C plays training-split Wordle with the current checkpoint, branches at early decision points (factor~2, rounds~1--3, $T{=}0.7$), and pairs the immediate diverging responses from the highest- and lowest-scoring siblings. DPO against a frozen reference ($\beta{=}0.20$) targets the residual distribution gap.
Branch-DPO follows a 3.0-point increase in the submitted-model trajectory, whereas the controlled single-host comparison shows a 2.07-point clemscore decrease and a 1.23-point statscore increase (Tables~\ref{tab:trajectory} and~\ref{tab:ablation}). We therefore report its contribution as evaluation-rail-dependent while retaining it as part of the submitted model's training sequence.
Appendix~\ref{app:branch-dpo} details instance selection, branch grouping, downstream scoring, and pair extraction.

\subsection{Preserving General Capability}
\label{sec:scaling}
By this point, every strong interactive candidate pays a specialization tax: statscore drops several points from the starting checkpoint on our pinned development set. Rather than add more data or replay, we treat preservation as a post-training weight-space problem. We scale the learned delta, $W(s)=W_{\text{base}}+s\,\Delta_{\text{LoRA}}$, and select the scalar $s\in[0,1]$ by constrained model selection,
\begin{equation}
\begin{aligned}
s^\star &= \arg\max_{s}\ S\!\big(W(s)\big)\\
\text{s.t.}\quad C\!\big(W(s)\big)&\ge C_{\min},
\end{aligned}
\label{eq:scale}
\end{equation}
where $C_{\min}$ is a paired interactive gate on the pinned public-development set. At $s^\star{=}0.85$, the mean of two clemscore evaluations is 50.43 (Table~\ref{tab:trajectory}). The operation is training-free and composes with any plain LoRA adapter, but the scalar must be selected on both axes: too low a value fails the interactive gate. In the official final evaluation, the submitted fp32 merge scores 38.92 public clemscore with statscore 44.14 against the 2B baseline's 44.24.

\section{Results}
\label{sec:results}

\begin{table*}[tbp]
\centering\small
\setlength{\tabcolsep}{4pt}
\begin{tabular}{@{}llrrrrrr@{}}
\toprule
System & Base & Public & Stat & Closed ID & Closed OOD & $\Delta$ID & $\Delta$OOD \\
\midrule
\multirow{5}{*}{Baseline} & Qwen3.5-2B & 10.67 & 44.24 & 13.41 & 3.72 & --- & --- \\
 & Qwen3.5-4B & 26.66 & 51.33 & 34.02 & 17.99 & --- & --- \\
 & Llama-3.1-8B & 19.53 & 45.59 & 31.24 & 22.62 & --- & --- \\
 & Qwen3.5-9B & 31.91 & 53.90 & 41.12 & 24.91 & --- & --- \\
 & Qwen3.5-27B & 60.30 & 65.05 & 64.34 & 43.51 & --- & --- \\
\midrule
LLP llp-final & Qwen3.5-9B & 53.39 & 57.80 & 59.23 & 22.09 & $+18.11$ & $-2.82$ \\
DAIR sft-dpo-v2 & Qwen3.5-2B & 49.60 & 43.53 & 50.75 & 14.56 & $+37.34$ & $+10.84$ \\
SLED-BSU & Qwen3.5-27B & 46.90 & 63.80 & 56.36 & 30.02 & $-7.98$ & $-13.49$ \\
CityUoL GuidePlay & Qwen3.5-2B & 46.66 & 42.30 & 46.26 & 10.25 & $+32.85$ & $+6.53$ \\
DAIR sft-v1 & Qwen3.5-2B & 46.01 & 44.35 & 46.57 & 15.62 & $+33.16$ & $+11.90$ \\
\textbf{playornotplay (ours)} & \textbf{Qwen3.5-2B} & \textbf{38.92} & \textbf{44.14} & \textbf{41.17} & \textbf{7.88} & $\mathbf{+27.76}$ & $\mathbf{+4.16}$ \\
BSU-SLIM & Qwen3.5-9B & 36.61 & 49.92 & 39.64 & 15.18 & $-1.48$ & $-9.73$ \\
Dialogue Architects & Qwen3.5-9B & 34.39 & 53.26 & 43.12 & 23.37 & $+2.00$ & $-1.54$ \\
Bentel iter\_3 & Qwen3.5-4B & 33.35 & 54.27 & 37.76 & 16.60 & $+3.74$ & $-1.39$ \\
Bentel iter & Qwen3.5-4B & 31.58 & 54.27 & 37.41 & 16.60 & $+3.39$ & $-1.39$ \\
Bentel iter\_2 & Qwen3.5-4B & 24.27 & 47.37 & 31.97 & 10.19 & $-2.05$ & $-7.80$ \\
\bottomrule
\end{tabular}
\caption{Official final evaluation results of the LM Playschool Challenge (\url{https://github.com/lm-playpen/lm-playschool-2026-final-results}, snapshot \texttt{2dd5a533}). Public is the public interactive clemscore; Stat is the static benchmark score; Closed ID and OOD are the organizer's hidden in-domain and out-of-domain clemscores. $\Delta$ID and $\Delta$OOD are differences against each system's own matching base. Submissions are sorted by public clemscore.}
\label{tab:main}
\end{table*}

Table~\ref{tab:main} reports the official final evaluation.
Our submitted artifact\footnote{The submitted checkpoint is the fp32-merged model \texttt{chnln/Qwen3.5-2B-playpen-playornotplay} at Hugging Face revision \texttt{828e356}. Table~\ref{tab:hparams} gives the complete training configuration; Appendix~\ref{app:rails} describes the evaluation protocols.} improves public clemscore from 10.67 to 38.92 ($+28.25$) and closed in-domain score from 13.41 to 41.17 ($+27.76$), both above the official 4B and 9B baselines. For comparison, scaling the official baseline from 2B to 9B raises public clemscore from 10.67 to 31.91 ($+21.24$) with no training at all.
Aggregate static performance is approximately preserved: 44.14 against the baseline's 44.24.

\begin{table}[t]
\centering\footnotesize
\setlength{\tabcolsep}{1.5pt}
\begin{tabular}{@{}lccc@{}}
\toprule
 & Played & Quality & Clem \\
 & (base $\to$ ours) & (base $\to$ ours) & (base $\to$ ours) \\
\midrule
Public & 27.22 $\to$ 78.63 & 39.20 $\to$ 49.50 & 10.67 $\to$ 38.92 \\
Closed ID & 47.32 $\to$ 82.62 & 28.33 $\to$ 49.83 & 13.41 $\to$ 41.17 \\
Closed OOD & 35.19 $\to$ \textbf{33.08} & 10.56 $\to$ 23.81 & 3.72 $\to$ 7.88 \\
\bottomrule
\end{tabular}
\caption{Played and Quality decomposition of the official final evaluation. OOD Played falls below the untrained base, indicating that the OOD clemscore gain is driven entirely by higher conditional Quality on the episodes that remain in play.}
\label{tab:decomp}
\end{table}

Table~\ref{tab:decomp} decomposes the official endpoint into Played and Quality. The in-domain gain is driven by both components: Played nearly doubles while Quality also increases. Out-of-domain Played, however, falls below the untrained base (33.08 vs.\ 35.19), so the OOD clemscore gain from 3.72 to 7.88 reflects only conditional quality on fewer played episodes.

Of the total positive out-of-domain movement (84.86 effective score), three held-out Wordle-crazy configurations contribute 70.02, or 82.5\%; the remainder comes from TA Frozen Lake ($+9.17$), TA Mastermind ($+5.00$), and Cryptolect ($+0.67$). The largest hidden in-domain gains are in games we never targeted: TMW GraphReasoning ($+71$), ImageGame ($+60$), and AdventureGame ($+58$), while the targeted Wordle family remains near zero on hidden in-domain episodes (3.3, 10.0, and 3.3). Tables~\ref{tab:official-id} and~\ref{tab:official-ood} give the full per-game decompositions.

On the pinned public-development set, the four-phase trajectory progresses from 13.05 through 43.85 (Phase~A), 46.83 and 45.52 (Phase~B passes), and 48.52 (Phase~C) to a mean 50.43 after delta scaling (Table~\ref{tab:trajectory}). Most adjacent-phase differences are small relative to the overall gain, and per-game changes outside Wordle are mixed (Tables~\ref{tab:app-pergame} and~\ref{tab:app-wordle}), so we report these as a trajectory rather than a decomposition into causal effects.

\section{Discussion}
\label{sec:discussion}

The official evaluation combines a large in-domain gain with weak out-of-domain transfer. The training trajectory and its follow-up experiments point to a common pattern: supervision is most useful when matched to what the current policy still lacks. Broad imitation creates participation, local preferences repair mechanically identifiable decisions, and weight-space selection chooses an operating point that retains the resulting gains.

\subsection{Acquisition Creates the Conditions for Local Repair}

Phase-level evidence comes from the pinned public-development trajectory and the controlled single-host campaign (Appendix~\ref{app:ablation}). Phase~A is the largest single step in that trajectory (Table~\ref{tab:trajectory}). Played rises sharply while conditional Quality also improves (Appendix~\ref{app:pergame}), and the final model's official decomposition is consistent with this breadth: its largest hidden in-domain gains appear in games no later phase targeted (TMW GraphReasoning, ImageGame, AdventureGame; Table~\ref{tab:official-id}). The breadth of these final-model gains is consistent with the role of the only all-game training stage.

The later phases operate in a narrower regime. Only turn-DPO pass~1 has a positive clem marginal on both the submitted-model trajectory and the controlled campaign ($+2.98$ and $+1.49$); pass~2 and branch-DPO are sign-inconsistent across the two rails (Tables~\ref{tab:trajectory} and~\ref{tab:ablation}). The clearest repair signal remains inside the Wordle family: three held-out Wordle-crazy configurations account for 82.5\% of total positive out-of-domain movement, while the targeted Wordle family remains near zero on hidden in-domain episodes (Tables~\ref{tab:official-ood} and~\ref{tab:official-id}). The observed transfer is therefore concentrated in related Wordle variants.

Locality appears to be the hinge. In an earlier experiment, the preference covered the entire remaining dialogue rather than the next post-feedback action. Unfiltered whole-dialogue DPO scored 12.35 and produced natural-language preambles that violated game protocols across Wordle, Codenames, and other games; filtering the chosen side to stronger source models partially recovered the score to 25.76 but left the principal abort patterns intact (Appendix~\ref{app:abandoned}). Taken together, the positive turn-local result and the whole-dialogue failures suggest that the clearest preference signal arises when the contrast is attached to a mechanically verifiable post-feedback decision. Whole-dialogue DPO, which paired entire remaining interactions rather than single responses, did not preserve protocol behavior even after source filtering.

\subsection{Preservation as Weight-Space Model Selection}

Delta scaling addresses a different problem from acquisition and repair. It adds no supervision and takes no gradient step; instead, it selects a point along the learned LoRA update. In the official evaluation, aggregate static performance is approximately preserved (44.14 against the baseline's 44.24), while the interactive gains remain substantial. The aggregate nevertheless hides redistribution: BBH Quality rises from 0.00 to 30.47, whereas IFEval falls from 68.52 to 51.85 and EQ-Bench Quality from 64.78 to 56.43.

Preservation is therefore a model-selection problem along the learned update, rather than another supervision problem. Delta scaling selects an operating point between interactive gain and aggregate static performance, with component-level redistribution.

\subsection{Training Effects Depend on Stage and Parent}

To test whether correction-based preference training remained useful after model selection, we continued DPO from the selected $s{=}0.85$ parent using solver corrections, success-versus-failure pairs, or teacher corrections. None of the five arms exceeded the 50.43 development endpoint; their scores ranged from 44.15 to 48.92 (Appendix~\ref{app:abandoned}). At this point in the pipeline, additional correction-based training did not produce a stronger model.

The absence of further gains does not mean that the 2B base had exhausted its headroom. DAIR's SFT-only system uses the same Qwen3.5-2B base and reaches 46.01 public, 46.57 hidden in-domain, and 15.62 hidden out-of-domain clemscore. This same-base comparison points back to acquisition: stronger broad supervision can create a better starting policy before local repair begins.

The effect of an acquisition update also depends on its parent. In a matched 4B probe, low-dose SFT on 700 off-policy successful episodes reduced clemscore from 40.75 to 24.36 on the same host, dataset revision, and evaluation rail. A separate 4B probe, trained on 303 self-generated successful dialogues, reached 35.06 (Appendix~\ref{app:abandoned}). Together with the late-continuation results, these probes suggest that training effects are path-dependent: broad imitation helps when participation is missing, local preference learning helps once failures become mechanically identifiable, and further updates after the policy or parent has changed can instead degrade behavior the model already possesses.

\subsection{Model Selection Shapes What Generalizes}

Across the eleven submissions, public clemscore is strongly associated with closed in-domain score ($r{=}0.93$) but only weakly with closed out-of-domain score ($r{=}0.30$). Every decision in our pipeline was made on the public-development set, and the final result mirrors that selection signal: a large public and hidden in-domain gain, but an out-of-domain clemscore of 7.88 and out-of-domain Played below the untrained base. The public set was informative about hidden in-domain progress but offered little guidance for broad transfer.

This pattern exposes a practical limit of the recipe-development process: the endpoint combines a large improvement over a weak starting point with limited absolute out-of-domain capability. Broader transfer would require either diagnostics or a selection signal that reaches beyond the Wordle family and the public development distribution.

\section{Conclusion}

Interactive dialogue-game failures can reflect both broad capability deficits and local decision errors. In our pipeline, broad SFT produces the largest improvement, while turn-local preference pairs provide the clearest positive repair signal among the preference-tuning variants we tested.
Combining broad SFT, turn-local repair, and delta scaling raises official public clemscore from 10.67 to 38.92 and closed in-domain score from 13.41 to 41.17, above the official 4B and 9B baselines, while approximately preserving aggregate static performance.
Out-of-domain transfer remains limited: clemscore is 7.88 and Played falls below the untrained base. The results support a scoped recipe for acquiring in-domain game participation and repairing precise within-family failures, with transfer concentrated in related Wordle variants.

\section*{Limitations}

\paragraph{Scope of mechanically verifiable repair.}
Turn-local repair relies on game-specific diagnostics that verify whether a post-feedback action violates an explicit state constraint. That precision restricts demonstrated coverage to the Wordle family, the only game for which Phases~B and~C construct preference data; the method has not been tested on failures requiring semantic or strategic judgment. The demonstrated out-of-domain coverage is related-variant transfer: three held-out Wordle-crazy configurations account for 70.02 of 84.86 total positive movement (82.5\%). The submitted artifact uses a single Qwen3.5-2B base, so scale applicability remains untested.

\paragraph{Untested dependence on the acquisition-stage parent.}
Broad SFT accounts for most of the measured improvement before repair is applied. The repair phases are evaluated after one specific, capped broad-SFT recipe; the study does not vary the strength or coverage of the acquisition-stage parent while holding the repair corpus and training configuration fixed. The experiments therefore cannot establish whether turn-local repair complements a stronger imitation stage, compensates for weaknesses specific to the submitted parent, or becomes less useful after stronger acquisition. Local repair is informative in the regime this parent creates; its contribution under a differently trained parent remains unknown.

\paragraph{Limits of phase attribution and model selection.}
The official evaluation validates the submitted endpoint, but the sequential trajectory does not identify the causal contribution of each phase. Phase and scaling choices reuse a 67-episode public-development set, and later-phase marginals are not consistently signed across the two evaluation rails. Two later phases are sign-inconsistent across rails: turn-DPO pass~2 is $-1.31$ on the development trajectory and $+1.34$ on the controlled campaign; branch-DPO is $+3.00$ and $-2.07$. Only pass~1 is positive on both ($+2.98$ and $+1.49$). Adjacent checkpoint differences combine training, sequential dependence, repeated selection, and evaluation variance, so they describe a trajectory rather than additive causal effects. The endpoint improves substantially over the base; the magnitude and necessity of individual later phases remain uncertain.

\paragraph{Aggregate rather than component-wise preservation.}
Delta scaling approximately preserves the official aggregate statscore (44.14 vs.\ 44.24) while retaining interactive gains. The preservation objective is itself aggregate: Eq.~\ref{eq:scale} maximizes aggregate static score subject to an aggregate interactive gate, so gains in one component can offset losses in another by design. BBH Quality rises from 0.00 to 30.47 while IFEval drops from 68.52 to 51.85 and EQ-Bench Quality from 64.78 to 56.43. The defensible conclusion is aggregate-level trade-off control, not preservation of every general capability. A stronger claim would require component-wise floors or a multi-objective selection criterion.

\section*{Acknowledgments}

We appreciate the helpful comments and suggestions from the anonymous reviewers.
This work is funded by the Dutch Research Council (NWO) through the AiNed Fellowship Grant NGF.1607.22.002, \textit{Dealing with Meaning Variation in NLP}.
This work used the Dutch national e-infrastructure with the support of the SURF Cooperative using grant no. EINF-18781/L1.

\bibliography{references}

\appendix
\newpage

\section{Training and Evaluation Details}
\label{app:training}

\subsection{Submitted Model}
\label{app:artifact}

The submitted artifact is the checkpoint at Hugging Face revision \texttt{828e356}~\footnote{\url{https://huggingface.co/chnln/Qwen3.5-2B-playpen-playornotplay}}. This is the revision used for the official evaluation.

The submitted model is an fp32 merge of a LoRA adapter into the Qwen3.5-2B base. Its training costs approximately 5.3 hours on a single A100-80GB GPU.

\subsection{Evaluation Pins}
\label{app:rails}

The paper reports scores on two dev/test sets, which are not directly comparable.

\paragraph{Official final evaluation.}
The organizer's evaluation at snapshot \texttt{2dd5a533} uses clemcore version 3.7.2. The official metadata does not record the exact clembench game-tree revision. Public, closed in-domain, and closed out-of-domain scores constitute the primary result. The public suite contains 502 episodes (72 interactive, 430 static); the closed suites contain 1{,}272 in-domain and 360 out-of-domain episodes. All runs use temperature 0, a 5{,}000-token generation limit, and thinking mode disabled.

\paragraph{Pinned public-development evaluation.}
Our development evaluations pin dataset revision \texttt{557d8caf} and clembench tree commit \texttt{ed39486}, yielding 67 interactive episodes across 14 games. This set produces the per-phase trajectory in Table~\ref{tab:trajectory} and the matched per-game comparison in Table~\ref{tab:app-pergame}.

\subsection{Training Details}
\label{app:hparams}

\subsubsection{Toolkits}

The training pipeline uses Playpen trainers built on Hugging Face \texttt{transformers}, \texttt{peft}, and \texttt{trl}. Phase~A uses TRL's \texttt{SFTTrainer}; Phases~B and~C use \texttt{DPOTrainer}.

\begin{table*}[t]
\centering\footnotesize
\setlength{\tabcolsep}{4pt}
\begin{tabular}{@{}p{.03\textwidth}p{.14\textwidth}p{.11\textwidth}p{.13\textwidth}p{.21\textwidth}p{.26\textwidth}@{}}
\toprule
Ph. & Data & Objective & LR / sched.\ / warmup & Batch / length / epochs & Other \\
\midrule
A & 9{,}522 rows & SFT, AdamW & 2e-4 / cosine / 0.03 & micro 8 $\times$ accum 4 $=$ 32; 2{,}048; 1 & LoRA $r$16/$\alpha$32/dropout 0.05/all-linear; bf16; grad.\ ckpt.\ on; seed 42 \\
\addlinespace
B1 & 1{,}905 pairs & DPO, $\beta{=}0.1$ & 5e-6 / cosine / 0.03 & micro 1 $\times$ accum 16 $=$ 16; 1{,}024; 1 & bf16; grad.\ ckpt.\ off; separate frozen reference \\
\addlinespace
B2 & 1{,}906 pairs & DPO, $\beta{=}0.1$ & 5e-6 / cosine / 0.03 & micro 1 $\times$ accum 16 $=$ 16; 1{,}024; 1 & bf16; grad.\ ckpt.\ off; separate frozen reference \\
\addlinespace
C & 77 pairs & DPO, $\beta{=}0.2$ & 5e-6 / cosine / 0.03 & micro 1 $\times$ accum 16 $=$ 16; 1{,}024; 1 & bf16; grad.\ ckpt.\ on; separate frozen reference; factor 2, rounds 1--3, $T{=}0.7$, 256 new tokens \\
\addlinespace
D & --- & delta scaling $s{=}0.85$ & no optimizer & --- & multiply each \texttt{lora\_B} by 0.85, then merge in fp32 \\
\bottomrule
\end{tabular}
\caption{Complete training configuration for the four effective phases. All trainable phases continue the same LoRA adapter on Qwen3.5-2B in bf16.}
\label{tab:hparams}
\end{table*}

\subsubsection{Broad-SFT Corpus}
\label{app:sft-data}

Phase~A trains on success-only episodes from \texttt{playpen-data}, capped at 700 per game, with seed-42 shuffling and a 95/5 train--eval split.
Table~\ref{tab:sft-games} lists per-game counts at the training-time snapshot.

\begin{table}[t]
\centering\small
\setlength{\tabcolsep}{4pt}
\begin{tabular}{@{}lrrr@{}}
\toprule
Game & Success & Capped & Train \\
\midrule
AdventureGame & 429 & 429 & 403 \\
Codenames & 2{,}402 & 700 & 669 \\
DoND & 1{,}256 & 700 & 663 \\
GuessWhat & 2{,}170 & 700 & 662 \\
HotAirBalloon & 1{,}102 & 700 & 658 \\
ImageGame & 1{,}018 & 700 & 662 \\
MatchIt ASCII & 1{,}654 & 700 & 668 \\
PrivateShared & 228 & 228 & 221 \\
ReferenceGame & 3{,}872 & 700 & 658 \\
Taboo & 2{,}522 & 700 & 678 \\
TextMapWorld & 742 & 700 & 666 \\
TMW GraphReasoning & 356 & 356 & 338 \\
TMW SpecificRoom & 717 & 700 & 660 \\
Wordle & 371 & 371 & 351 \\
Wordle w/ Clue & 438 & 438 & 419 \\
Wordle w/ Critic & 740 & 700 & 669 \\
\midrule
\textbf{Total} & \textbf{20{,}017} & \textbf{9{,}522} & \textbf{9{,}045} \\
\bottomrule
\end{tabular}
\caption{Broad-SFT per-game counts at the reconstructed training-time snapshot (\texttt{cf23af92}). Success counts successful episodes; Capped applies the 700-per-game limit; Train is the count after the 95/5 split.}
\label{tab:sft-games}
\end{table}

\subsubsection{Turn-Local Preference Construction}
\label{app:turn-dpo}

Phase~B constructs preference pairs from clean successful Wordle episodes. Eligibility is decided in three steps:

\begin{enumerate}
\itemsep0pt
\item The source row has game \texttt{wordle} and outcome \texttt{success}. The optional source-model filter was empty, so all successful episodes contribute.
\item The whole episode is rejected if any user message contains ``not a valid word'' or if any parsed assistant guess repeats within the episode.
\item The selected assistant reply contains a parseable lowercase \texttt{guess:}~field, is assistant turn~2 or later, and immediately follows a user message containing \texttt{guess\_feedback:}.
\end{enumerate}

The prompt is the exact dialogue history before that reply. The preferred completion $y^+$ is the unmodified response from the clean successful episode; it is not a teacher-generated correction. The rejected completion $y^-$ applies one of four deterministic corruptions to that same response:

\begin{itemize}
\itemsep0pt
\item \textbf{repeat\_last}: replace the current guess with the previous guess; skip if already equal.
\item \textbf{bad\_length}: replace a guess of at least five characters with its first four.
\item \textbf{invalid\_word}: try fixed single-letter substitutions and then fixed candidates, taking the first five-letter alphabetic token absent from the official Wordle lexicon, with \texttt{zzzzz} as fallback.
\item \textbf{duplicate\_guess\_keyword}: append a second \texttt{guess:} line to the otherwise unchanged response.
\end{itemize}

Pass~1 uses repeat\_last and bad\_length; pass~2 uses invalid\_word and duplicate\_guess\_keyword. One-negative sampling was not enabled, so every configured corruption that applies produces a pair.
Table~\ref{tab:wordle-real} shows a real failure of the kind the corruptions imitate.

\begin{table}[t]
\centering\small
\setlength{\tabcolsep}{3pt}
\begin{tabular}{@{}p{.16\linewidth}p{.76\linewidth}@{}}
\toprule
Role & Content \\
\midrule
Model & \texttt{guess: slate} \\
GM & \texttt{guess\_feedback:} s\textless red\textgreater{} l\textless red\textgreater{} a\textless red\textgreater{} t\textless red\textgreater{} e\textless red\textgreater \\
Model & \texttt{guess: crane} \\
GM & \texttt{guess\_feedback:} c\textless red\textgreater{} r\textless green\textgreater{} a\textless red\textgreater{} n\textless green\textgreater{} e\textless red\textgreater \\
Model & The feedback shows that `r' and `n' are correct in their positions, while `c', `a', and `e' are incorrect.\ \ldots\ \texttt{guess: ringer} \\
GM & \texttt{Error: The length of the guessed word is not 5.} \texttt{game\_result = ABORT} \\
\bottomrule
\end{tabular}
\caption{A real Wordle failure from the broad-SFT checkpoint on the pinned public-development set (medium-frequency instance~14, target word \emph{grind}). The model reasons correctly about letter positions and then produces a six-letter guess. This is a diagnostic example, not a Phase~B training row: Phase~B pairs are built from clean successful episodes.}
\label{tab:wordle-real}
\end{table}

\subsubsection{Branch-DPO Construction}
\label{app:branch-dpo}

Phase~C loads the train-split instance configuration, filters to Wordle, shuffles with seed~42, and selects 27 instances. The current checkpoint plays each instance with branching factor~2 at rounds~1--3, temperature 0.7, and at most 256 generated tokens per turn. Sibling branches are grouped by shared game-snapshot origin, and a group needs at least two branches. Within a group, siblings are ranked by the score of the complete downstream episode, which combines terminal outcome, speed, closeness, strategy, and penalties for repeated, invalid, bad-length, duplicated, missing, or feedback-inconsistent guesses; it is not a binary win/loss label. The chosen and rejected items are only the immediate diverging assistant responses of the highest- and lowest-scoring siblings, with the shared history as prompt. A group is dropped when the two scores are equal.

The run produced 90 trajectories, 295 branch points, and 77 DPO pairs (73 train, 4 eval).

\subsection{Two Nominal Stages Removed}
\label{app:noop}

The training process for the submitted model was originally described as six stages. Nominal Stage~2 was a continued SFT pass over the weakest games; nominal Stage~3 was a GRPO~\citep{shao2024deepseekmath} probe on Wordle with a dense reward that parses feedback and penalizes six mechanical violations. Neither changed a single weight, for two unrelated reasons.

\paragraph{Nominal Stage~2.}
The run loaded the broad-SFT adapter through the inference loader, which sets \texttt{is\_trainable=False}, and the warm-start path passed \texttt{peft\_config=None}, so no new trainable adapter was created. TRL's \texttt{enable\_input\_require\_grads()} kept the forward and backward passes from raising, so the run completed and wrote checkpoints. The output adapter is byte-identical to the broad-SFT checkpoint (372 of 372 tensors equal) and its optimizer state contains zero parameters.

\paragraph{Nominal Stage~3.}
With $\beta \neq 0$, \texttt{GRPOTrainer} in TRL 0.29.1 builds the frozen reference by calling \texttt{add\_adapter} with the same configuration object, which still carries \texttt{inference\_mode=True}. Adapter injection therefore silently re-froze every adapter, the policy included, and the optimizer was constructed afterwards on an empty parameter set. The log reports 16.8M trainable parameters and 54 non-zero gradient norms, but every gradient was computed and discarded; the KL term is identically zero across all 54 steps.

\paragraph{Behavioral corroboration.}
Re-evaluating the three nominal checkpoints at temperature~0 on the pinned split, 13 of 14 games produce byte-identical transcripts. The one divergence is AdventureGame, traced to Python set iteration order under \texttt{PYTHONHASHSEED}. The three scores 43.85, 44.09, and 43.35 are thus three evaluations of one set of weights, and their 0.74-point spread is a direct measurement of evaluation noise on this suite.

\paragraph{Corrected re-runs.}
Both stages were re-run with the bugs fixed on a single host, using the same seeds, learning rates, batch geometry, and reward shaping as the originals. Because the corrected runs used the paper's pinned dataset revision rather than the original snapshot, the weak-game SFT corpus changed from 2{,}119 to 2{,}224 rows and the GRPO instance set shifted by two task IDs. The corrected runs therefore reproduce the recipe family with slightly different corpora. The corrected weak-game SFT scored 42.07 ($-1.78$ from its 43.85 parent) and the corrected GRPO scored 43.55 ($+1.48$ from that parent), a net $-0.30$ over both. Training dynamics confirm genuine optimization: non-zero gradient norms where the originals showed zeros, and a non-zero KL term where the original was identically zero.

\subsection{Controlled Single-Host Ablation}
\label{app:ablation}

Table~\ref{tab:ablation} reports a controlled campaign run on one host, approximately one month before the final Hub commit. All rows share the same host, harness, and session. The campaign predates the submission and was available at ship time.

\begin{table}[t]
\centering\small
\setlength{\tabcolsep}{4pt}
\begin{tabular}{@{}llrr@{}}
\toprule
Row & Configuration & Clem & Stat \\
\midrule
A & broad SFT only & 44.09 & 39.88 \\
B & A $+$ turn-DPO pass~1 & 45.58 & 39.62 \\
C & A $+$ turn-DPO pass~1 (repeat) & 45.94 & 40.96 \\
D & C $+$ turn-DPO pass~2 & 47.28 & 40.31 \\
E & D $+$ branch-DPO & 45.21 & 41.54 \\
\bottomrule
\end{tabular}
\caption{Controlled single-host ablation. Rows~B and~C independently apply the same turn-DPO pass-1 configuration to the same effective parent; their difference ($+0.36$ clem, $+1.34$ stat) estimates repeat-run variation. Adding branch-DPO to Row~D reduces clem by 2.07 and raises stat by 1.23 (Row~E). Row~A's adapter is byte-identical to the broad-SFT adapter that scored 43.85 on the pinned rail; the 0.24-point difference is evaluation variation measured in \S\ref{app:noop}. This controlled campaign and Table~\ref{tab:trajectory} use separate evaluation rails.}
\label{tab:ablation}
\end{table}

Of the three later phases, only turn-DPO pass~1 has a positive clem marginal on both the submitted-model trajectory and this campaign ($+2.98$ and $+1.49$); pass~2 ($-1.31$/$+1.34$) and branch-DPO ($+3.00$/$-2.07$) are sign-inconsistent across the two rails. A consistent trade appears across phases: the phases that raise clem cost a small amount of stat, and the one that costs clem gains stat.

\subsection{Abandoned Directions}
\label{app:abandoned}

\paragraph{Late continuation.}
Table~\ref{tab:late-cont} reports ten evaluated endpoints from the late-continuation campaign: eight DPO runs and two post-DPO training-free rescalings.

\begin{table}[t]
\centering\small
\setlength{\tabcolsep}{4pt}
\begin{tabular}{@{}lrrr@{}}
\toprule
Arm & Pairs & Clem & vs.\ 50.43 \\
\midrule
Wordle-only solver & 66 & 48.39 & $-2.0$ \\
All failures & 675 & 44.15 & $-6.3$ \\
Wordle high-dose & 66 & 44.99 & $-5.4$ \\
Conservative SVF & 500 & 48.92 & $-1.5$ \\
SVF, then $s{=}0.95$ & --- & 48.08 & $-2.4$ \\
SVF, then $s{=}0.90$ & --- & 46.68 & $-3.8$ \\
SVF from Phase~A parent & 500 & 49.76 & $-0.7$ \\
SVF from unscaled Phase~C & 500 & 48.74 & $-1.7$ \\
GPT-5.5 teacher, shipped parent & 411 & 46.35 & $-4.1$ \\
GPT-5.5 teacher, Phase~A parent & 411 & 27.20 & $-23.2$ \\
\bottomrule
\end{tabular}
\caption{Ten evaluated late-continuation endpoints. Eight are DPO runs ($\beta{=}0.20$, max length 1{,}024, micro-batch 1, separate frozen reference); two are training-free rescalings of a preceding DPO arm (marked ---). SVF denotes success-versus-failure pairs. Of the eight DPO runs, five start from the selected $s{=}0.85$ adapter and three start from other parents (Phase~A or unscaled Phase~C). The 50.43 comparator is the mean of two evaluations of the selected adapter and serves as the contextual reference for this campaign.}
\label{tab:late-cont}
\end{table}

Solver pairs identify a bad Wordle turn and pair the actual response against a rule-based correction. Success-versus-failure pairs select the first error-followed assistant turn, or otherwise the last, and match a successful same-game response by prompt similarity. The teacher corpus is 345 GPT-5.5 corrections plus 66 solver corrections. Continuation of an existing adapter passes no new PEFT configuration, so the effective adapter architecture remains the parent's, despite what some launcher variables advertise.

Both cross-parent comparisons in the table jointly vary parent, learning rate, and epoch count (arms 4 vs.\ 7 for the 500-pair SVF corpus; arms 9 vs.\ 10 for the 411-pair teacher corpus).

\paragraph{Whole-trajectory preference tuning.}
Two earlier DPO arms trained on full-dialogue preference pairs from the broad-SFT checkpoint, continuing at LR 5e-6, $\beta{=}0.1$, micro-batch~1 $\times$ accumulation~8, max length 2{,}048, one epoch. The prompt was the initial user message and both completions were entire remaining dialogues, pairing one success against one random failure or abort from the same game, experiment, task, role, and opening message.

\begin{itemize}
\itemsep0pt
\item \textbf{Unfiltered whole-dialogue DPO}: the chosen side is any successful transcript; 3{,}973 pairs. Clemscore 12.35. The model began emitting natural-language preambles before game actions, violating strict request format across Wordle, Codenames, and other games.
\item \textbf{Source-filtered whole-dialogue DPO}: the chosen side is filtered to transcripts from six stronger source models; 1{,}337 pairs. Clemscore 25.76. Protocol compliance partially recovered, but Wordle and Codenames still aborted; all six GuessWhat episodes were played, but none was won.
\end{itemize}

No live teacher was queried in either arm: the source-filtered variant uses transcripts that already exist in \texttt{playpen-data}, filtering them by source-model name. Unfiltered whole-dialogue preference learning did not merely fail to help; it destroyed protocol compliance across multiple games. The contrast with Phase~B motivates held-fixed histories and token-level contrasts.

\paragraph{4B supervised probes.}
We tested two SFT-only updates on Qwen3.5-4B. In the matched off-policy probe, a fresh LoRA adapter trained for one epoch on 700 successful episodes reduced clemscore from 40.75 to 24.36 on the same host, dataset revision, clembench revision, and evaluation rail. A separate probe let the vanilla 4B model generate training trajectories, retained its 303 successful full dialogues across 14 games, and applied success-only SFT; the resulting model scored 35.06. Both probes used LR 2e-5, effective batch size 16, and maximum length 2{,}048.

\subsection{Official Per-Game Decompositions}
\label{app:official-pergame}

Tables~\ref{tab:official-id} and~\ref{tab:official-ood} decompose the official final evaluation per game for the 2B baseline and our submission. Table~\ref{tab:official-static} gives the public static components behind the reported statscores.

\begin{table}[t]
\centering\small
\setlength{\tabcolsep}{4pt}
\begin{tabular}{@{}lrrrr@{}}
\toprule
& \multicolumn{2}{c}{Baseline} & \multicolumn{2}{c}{Ours} \\
\cmidrule(lr){2-3}\cmidrule(lr){4-5}
Benchmark & Played & Quality & Played & Quality \\
\midrule
BBH & 100.00 & 0.00 & 100.00 & 30.47 \\
CLadder & 100.00 & 48.51 & 100.00 & 46.53 \\
EQ-Bench & 97.06 & 64.78 & 94.12 & 56.43 \\
IFEval & 100.00 & 68.52 & 100.00 & 51.85 \\
MMLU-Pro & 100.00 & 40.71 & 100.00 & 38.05 \\
\bottomrule
\end{tabular}
\caption{Public static components of the official evaluation. Applying Eq.~\ref{eq:clemscore} gives statscore 44.24 for the baseline and 44.14 for our submission. The aggregate is approximately preserved, but the components are redistributed: BBH Quality rises by 30.47 while IFEval falls by 16.67 and EQ-Bench Quality by 8.35.}
\label{tab:official-static}
\end{table}

\begin{table*}[t]
\centering\footnotesize
\setlength{\tabcolsep}{5pt}
\begin{tabular}{@{}lrrrrrrr@{}}
\toprule
& \multicolumn{3}{c}{Qwen3.5-2B baseline} & \multicolumn{3}{c}{playornotplay (ours)} & \\
\cmidrule(lr){2-4}\cmidrule(lr){5-7}
Game & Played & Quality & Eff. & Played & Quality & Eff. & $\Delta$ Eff. \\
\midrule
TMW GraphReasoning & 0.00 & 0.00 & 0.00 & 100.00 & 70.99 & 70.99 & $+70.99$ \\
ImageGame & 80.00 & 12.03 & 9.62 & 97.50 & 71.46 & 69.67 & $+60.05$ \\
AdventureGame & 0.00 & 0.00 & 0.00 & 100.00 & 58.33 & 58.33 & $+58.33$ \\
TextMapWorld & 0.00 & 0.00 & 0.00 & 100.00 & 57.91 & 57.91 & $+57.91$ \\
TMW SpecificRoom & 0.00 & 0.00 & 0.00 & 100.00 & 53.33 & 53.33 & $+53.33$ \\
DoND & 30.00 & 0.00 & 0.00 & 100.00 & 39.75 & 39.75 & $+39.75$ \\
Taboo & 93.33 & 8.63 & 8.05 & 100.00 & 47.78 & 47.78 & $+39.73$ \\
BBH & 100.00 & 0.00 & 0.00 & 100.00 & 39.42 & 39.42 & $+39.42$ \\
PrivateShared & 0.00 & 0.00 & 0.00 & 56.00 & 38.69 & 21.67 & $+21.67$ \\
GuessWhat & 0.00 & 0.00 & 0.00 & 100.00 & 18.33 & 18.33 & $+18.33$ \\
Codenames & 0.00 & 0.00 & 0.00 & 46.92 & 29.51 & 13.85 & $+13.85$ \\
MatchIt ASCII & 100.00 & 75.00 & 75.00 & 100.00 & 85.00 & 85.00 & $+10.00$ \\
Wordle w/ Clue & 6.67 & 50.00 & 3.33 & 10.00 & 100.00 & 10.00 & $+6.67$ \\
MMLU-Pro & 100.00 & 23.00 & 23.00 & 100.00 & 28.00 & 28.00 & $+5.00$ \\
Wordle & 16.67 & 0.00 & 0.00 & 43.33 & 7.69 & 3.33 & $+3.33$ \\
Wordle w/ Critic & 3.33 & 0.00 & 0.00 & 3.33 & 100.00 & 3.33 & $+3.33$ \\
Clean Up & 66.67 & 3.17 & 2.11 & 88.89 & 4.85 & 4.31 & $+2.20$ \\
CLadder & 100.00 & 52.00 & 52.00 & 100.00 & 50.00 & 50.00 & $-2.00$ \\
ReferenceGame & 100.00 & 45.56 & 45.56 & 100.00 & 43.33 & 43.33 & $-2.23$ \\
IFEval & 100.00 & 64.00 & 64.00 & 100.00 & 50.00 & 50.00 & $-14.00$ \\
EQ-Bench & 97.00 & 63.20 & 61.30 & 89.00 & 52.09 & 46.36 & $-14.94$ \\
\bottomrule
\end{tabular}
\caption{Official hidden in-domain per-game decomposition, 21 scored configurations, sorted by change in effective score. Eff.\ is the displayed diagnostic $\mathrm{Played}\times\mathrm{Quality}/100$; these values are not averaged to reconstruct the suite clemscore (Eq.~\ref{eq:clemscore}).}
\label{tab:official-id}
\end{table*}

\begin{table*}[t]
\centering\footnotesize
\setlength{\tabcolsep}{5pt}
\begin{tabular}{@{}lrrrrrrr@{}}
\toprule
& \multicolumn{3}{c}{Qwen3.5-2B baseline} & \multicolumn{3}{c}{playornotplay (ours)} & \\
\cmidrule(lr){2-4}\cmidrule(lr){5-7}
Game & Played & Quality & Eff. & Played & Quality & Eff. & $\Delta$ Eff. \\
\midrule
Wordle-crazy w/ Critic & 6.67 & 0.00 & 0.00 & 36.67 & 90.91 & 33.34 & $+33.34$ \\
Wordle-crazy w/ Clue & 6.67 & 50.00 & 3.33 & 36.67 & 100.00 & 36.67 & $+33.34$ \\
TA Frozen Lake & 50.00 & 26.11 & 13.05 & 40.00 & 55.56 & 22.22 & $+9.17$ \\
TA Mastermind & 0.00 & 0.00 & 0.00 & 15.00 & 33.33 & 5.00 & $+5.00$ \\
Wordle-crazy & 3.33 & 0.00 & 0.00 & 80.00 & 4.17 & 3.34 & $+3.34$ \\
Cryptolect & 56.67 & 0.00 & 0.00 & 33.33 & 2.00 & 0.67 & $+0.67$ \\
Clockwork Courier & 0.00 & 0.00 & 0.00 & 56.67 & 0.00 & 0.00 & $+0.00$ \\
Get to the Point & 40.00 & 0.00 & 0.00 & 35.00 & 0.00 & 0.00 & $+0.00$ \\
TA Sokoban & 10.00 & 0.00 & 0.00 & 10.00 & 0.00 & 0.00 & $+0.00$ \\
Tower of Hanoi & 20.83 & 0.00 & 0.00 & 25.00 & 0.00 & 0.00 & $+0.00$ \\
Chronicle & 100.00 & 5.02 & 5.02 & 3.33 & 0.00 & 0.00 & $-5.02$ \\
ST Clean Up & 83.33 & 11.65 & 9.71 & 33.33 & 4.85 & 1.62 & $-8.09$ \\
TA Blackjack & 80.00 & 23.33 & 18.66 & 25.00 & 18.67 & 4.67 & $-14.00$ \\
\bottomrule
\end{tabular}
\caption{Official hidden out-of-domain per-game decomposition, 13 scored configurations from eight game families, sorted by change in effective score. The three Wordle-crazy configurations contribute 82.5\% of total positive movement. Columns are as in Table~\ref{tab:official-id}.}
\label{tab:official-ood}
\end{table*}

\section{Per-Game Matched Comparison}
\label{app:pergame}

Table~\ref{tab:app-pergame} decomposes the pipeline per game at four checkpoints: the vanilla base, the broad-SFT checkpoint (Phase~A), the last Wordle-trained checkpoint (Phase~C, before delta scaling), and the shipped fp32 merge. To make the per-game deltas internally comparable, all four checkpoints were re-evaluated on a single host with the same evaluation harness, dataset revision \texttt{557d8caf}, and clembench commit \texttt{ed39486}; the matched base score coincides with the organizer-reported baseline (13.05). We use these matched runs for the per-game analysis and the official evaluations for the headline results in the body.

At suite level, macro-averaged Played rises from 30.95 to 84.19 between the base and Phase~A, while macro-averaged Quality over played games rises from 42.15 to 52.09.

The three delta columns separate what changed during broad SFT (Base$\to$SFT), during the Wordle-only phases (SFT$\to$Branch), and under gradient-free delta scaling plus the fp32 merge (Branch$\to$Final). Changes outside Wordle during the Wordle-only phases are mixed in sign and rest on 3--13 episodes per game; their interpretation is descriptive.

\begin{table*}[tbp]
\centering\small
\setlength{\tabcolsep}{4.5pt}
\begin{tabular}{@{}lrrrrrrrr@{}}
\toprule
Game & $n$ & Base & Broad SFT & Branch-DPO & Final & \shortstack{$\Delta$\\Base$\to$SFT} & \shortstack{$\Delta$\\SFT$\to$Branch} & \shortstack{$\Delta$\\Branch$\to$Final} \\
\midrule
AdventureGame & 5 & 0.0 & 33.3 & 40.0 & 33.3 & $+33.3$ & $+6.7$ & $-6.7$ \\
Codenames & 13 & 0.0 & 0.0 & 7.7 & 7.7 & $0.0$ & $+7.7$ & $0.0$ \\
GuessWhat & 6 & 0.0 & 50.0 & 33.3 & 50.0 & $+50.0$ & $-16.7$ & $+16.7$ \\
ImageGame & 4 & 35.8 & 81.2 & 89.8 & 85.2 & $+45.4$ & $+8.6$ & $-4.6$ \\
MatchIt ASCII & 4 & 75.0 & 50.0 & 75.0 & 75.0 & $-25.0$ & $+25.0$ & $0.0$ \\
PrivateShared & 4 & 0.0 & 35.1 & 21.3 & 18.3 & $+35.1$ & $-13.8$ & $-3.0$ \\
ReferenceGame & 5 & 100.0 & 80.0 & 60.0 & 80.0 & $-20.0$ & $-20.0$ & $+20.0$ \\
Taboo & 6 & 0.0 & 41.7 & 41.7 & 25.0 & $+41.7$ & $0.0$ & $-16.7$ \\
TextMapWorld & 5 & 0.0 & 55.6 & 65.2 & 65.7 & $+55.6$ & $+9.6$ & $+0.5$ \\
TMW GraphReasoning & 3 & 0.0 & 58.9 & 58.9 & 58.9 & $+58.9$ & $0.0$ & $0.0$ \\
TMW SpecificRoom & 3 & 0.0 & 66.7 & 66.7 & 66.7 & $+66.7$ & $0.0$ & $0.0$ \\
Wordle & 3 & 0.0 & 0.0 & 0.0 & 0.0 & $0.0$ & $0.0$ & $0.0$ \\
Wordle w/ Clue & 3 & 0.0 & 33.3 & 50.0 & 50.0 & $+33.3$ & $+16.7$ & $0.0$ \\
Wordle w/ Critic & 3 & 0.0 & 66.7 & 66.7 & 66.7 & $+66.7$ & $0.0$ & $0.0$ \\
\midrule
\textbf{Overall} & 67 & \textbf{13.05} & \textbf{43.85} & \textbf{48.52} & \textbf{48.70} & $+30.80$ & $+4.67$ & $+0.18$ \\
\bottomrule
\end{tabular}
\caption{Matched per-game comparison of four checkpoints on the pinned public-development set (single host, single harness, dataset \texttt{557d8caf}, clembench \texttt{ed39486}). The per-game score is the derived $P_g Q_g/100$; it is displayed as 0 when Played $=0$, although Quality itself is undefined there (Eq.~\ref{eq:clemscore}). The overall clemscore is the product of macro-averaged Played and macro-averaged Quality over played games, not the mean of the per-game column, so per-game deltas do not sum to the overall delta. The Broad SFT column is the Phase~A checkpoint and the Branch-DPO column is the Phase~C checkpoint (Table~\ref{tab:trajectory}). $\Delta$ Branch$\to$Final includes gradient-free delta scaling and the fp32 merge, not additional training. Each game has 3--13 episodes; the changes are descriptive.}
\label{tab:app-pergame}
\end{table*}

\section{Plain-Wordle Behavior by Phase}
\label{app:wordle}

Table~\ref{tab:app-wordle} tracks the three public plain-Wordle episodes across the pipeline, using process-level metrics rather than the game score alone. Among checkpoints with played plain-Wordle episodes, mean repetitions fall from 3.0 after broad SFT to 0 after the second Phase~B pass; repetition is unobservable after the first Phase~B pass and after scaling because no episode is played. Phase~C shows the most stable protocol behavior among the evaluated checkpoints (100\% played, no violated requests, 100\% request success), yet conditional Quality is 0 at every checkpoint with played episodes: the Wordle-focused phases repaired protocol-level failures on this public probe but did not solve the underlying guessing task.

Across repeated same-harness evaluations, the unscaled Phase~C checkpoint plays all three public episodes, while the $s{=}0.85$-scaled checkpoint and all downstream artifacts abort all three, re-introducing a bad-length guess of the kind targeted in the first Phase~B pass. Repeated evaluations establish this pattern on the fixed three-episode probe.

\begin{table*}[tbp]
\centering\small
\setlength{\tabcolsep}{3pt}
\begin{tabular}{@{}llrrrrrr@{}}
\toprule
Checkpoint & Training scope & Suite Clem & W Played & W Quality & Rep./played ep. & Viol./ep. & Req.\ success \\
\midrule
Broad SFT weights$^\ast$ & all games & 43.85 & 66.7 & 0 & 3.00 & 0.33 & 88.9 \\
Phase B pass 1 & Wordle & 46.83 & 0 & --- & --- & 1.00 & 63.9 \\
Phase B pass 2 & Wordle & 45.52 & 66.7 & 0 & 0.00 & 0.33 & 91.7 \\
Phase C & Wordle & 48.52 & 100 & 0 & 0.00 & 0.00 & 100.0 \\
$+$ scaling $s{=}.85$ & none (no gradient) & 50.82 / 50.03 & 0 & --- & --- & 1.00 & 69.4 \\
Shipped fp32 merge & none (merge only) & 48.70 / 49.10 & 0 & --- & --- & 1.00 & 69.4 \\
\bottomrule
\end{tabular}
\caption{Plain-Wordle behavior after each phase on the three public validation episodes (development evaluations; pinned dataset \texttt{557d8caf}). $^\ast$Three greedy evaluations of byte-identical weights span 43.35--44.09; the difference is evaluation noise (\S\ref{app:noop}). Suite Clem is the full 14-game clemscore; for the scaled and shipped artifacts we list the two runs of each. W Quality is undefined (---) when no episode is played. Rep./played ep.\ is the mean clembench \emph{Guess Repetitions} over played episodes; Viol./ep.\ and Req.\ success are the mean \emph{Violated Request Count} and \emph{Request Success Ratio} (\%) over all three episodes.}
\label{tab:app-wordle}
\end{table*}

\end{document}